\documentclass[11pt,letterpaper]{article}
\usepackage[T1]{fontenc}
\usepackage[utf8]{inputenc}
\usepackage[english]{babel}
\usepackage{emnlp2017}
\usepackage{times}
\usepackage{latexsym}
\usepackage{amssymb}
\usepackage{amsmath}
\usepackage{amsfonts}
\usepackage{tabularx}
\usepackage{booktabs}
\usepackage{graphicx}
\usepackage{url}

\emnlpfinalcopy

\def\emnlppaperid{686}

\title{Temporal dynamics of semantic relations in word embeddings:\\ an application to predicting armed conflict participants}

\author{Andrey Kutuzov \\
  Department of Informatics\\
  University of Oslo\\
  {\tt andreku@ifi.uio.no} \\ \And
  Erik Velldal \\
  Department of Informatics\\
  University of Oslo\\
  {\tt erikve@ifi.uio.no} \\ \And
  Lilja {\O}vrelid \\
  Department of Informatics\\
  University of Oslo\\
  {\tt liljao@ifi.uio.no}}

\date{}

\begin{document}
\maketitle
\begin{abstract}
This paper deals with using word embedding models to trace the temporal dynamics of semantic relations between pairs of words. The set-up is similar to the well-known analogies task, but expanded with a time dimension. To this end, we apply incremental updating of the models with new training texts, including incremental vocabulary expansion, coupled with learned transformation matrices that let us map between members of the relation.
The proposed approach is evaluated on the task of predicting insurgent armed groups based on geographical locations. The gold standard data for the time span 1994--2010 is extracted from the UCDP Armed Conflicts dataset. The results show that the method is feasible and outperforms the baselines, but also that important work still remains to be done. 
\end{abstract}

\section{Introduction and related work}
In this research, we make an attempt to model the dynamics of worldwide armed conflicts on the basis of English news texts. To this end, we employ the well-known framework of Continuous Bag-of-Words modeling \cite{Mikolov_representation:2013} for training word embeddings on the English Gigaword news text corpus \cite{Gigaword:11}. We learn linear projections from the embeddings of geographical locations where violent armed groups were active to the embeddings of these groups. These projections are then applied to the embeddings and gold standard data from the subsequent year, thus predicting what entities act as violent groups in the next time slice. To evaluate our approach, we adapt the UCDP Armed Conflict Dataset \citep{gleditsch2002armed,AllMelThe:17} (see Section~\ref{sec:gold} for details). 

Here is a simplified example of the task: given that in 2003, the \textit{Kashmir Liberation Front} and \textit{ULFA} were involved in armed conflicts in India, and \textit{Lord's Resistance Army} in Uganda, predict entities playing the same role in 2004 in Iraq (the correct answers are \textit{Ansar al-Islam}, \textit{al-Mahdi Army} and \textit{Islamic State}). The nature of the task is conceptually similar to that of analogical reasoning, but with the added complexity of temporal change.

Attempts to detect semantic change using unsupervised methods have a long history. Significant results have already been achieved in employing word embeddings to study diachronic language change. Among others, \citet{eger2016} show that the embedding of a given word for a given time period to a large extent is a linear combination of its embeddings for the previous time periods. 
\citet{hamilton2016cultural} proposed an important distinction between cultural shifts and linguistic drifts. They proved that global embedding-based measures (comparing the similarities of words to all other words in the lexicon) are sensitive to regular processes of linguistic drift, while local measures (comparing nearest neighbors' lists) are a better fit for more irregular cultural shifts in word meaning. 

Our focus here is on cultural shifts: it is not the dictionary meanings of the names denoting locations and armed groups that change, but rather their `image' in the analyzed texts. 
Our measurement approach can also be defined as `local' to some extent: the linear projections that we learn are mostly based and evaluated on the nearest neighborhood data. However, this method is different in that its scope is not single words but pairs of typed entities (`\textit{location}' and `\textit{armed group}' in our case) and the semantic relations between them.

\subsection{Contributions}
The main contributions of this paper are:
\begin{enumerate}
\item We show that distributional semantic models, in particular word embeddings, can be used not only to trace diachronic semantic shifts in words, but also the temporal dynamics of semantic relations between pairs of words. 
\item The necessary prerequisites for achieving decent performance in this task are incremental updating of the models with new textual data (instead of training from scratch each time new data is added) and some way of expanding the vocabulary of the models.
\end{enumerate}

\section{Gold standard data on armed conflicts}\label{sec:gold}
The UCDP/PRIO Armed Conflict Dataset maintained by the Uppsala Conflict Data Program and the Peace Research Institute Oslo is a manually annotated geographical and temporal dataset with information on armed conflicts, in the time period from 1946 to the present \citep{gleditsch2002armed,AllMelThe:17}. It encodes conflicts, where at least one party is the government of a state. The Armed Conflict Dataset is widely used in statistical and macro-level conflict research; however, it was adapted and introduced to the NLP field only recently, starting with \cite{kutuzov2017}. Whereas that work was focused on detecting the onset/endpoint of armed conflicts, the current paper further extends on this by using the dataset to evaluate the detection of changes in the semantic relation holding between participants of armed conflicts and their locations.

Two essential notions in the UCDP data are those of \textit{event} and \textit{armed conflict}. \textit{Events} can evolve into full-scale \textit{armed conflicts}, defined as contested incompatibilities that concern government and/or territory where the use of armed force between two parties, of which at least one is the government of a state, results in at least 25 battle-related deaths \cite{sundberg2013introducing}. 

The subset of the data that we employ is the \textit{UCDP Conflict Termination dataset} \cite{Kreutz:10}.
It contains entries on starting and ending dates of about 2000 conflicts. We limit ourselves to the conflicts taking place between 1994 and 2010 (the Gigaword time span). Almost always, the first actor of the conflict (\textit{sideA}) is the government of the corresponding location, and the second actor (\textit{sideB}) is some insurgent armed group we are interested in. We omitted the conflicts where both sides were governments (about 2\% of the entries) or where one of the sides was mentioned in the Gigaword less than 100 times (about 1\% of the entries). In cases when the UCDP described the conflict as featuring several groups on the \textit{sideB}, we created a separate entry for each. 

This resulted in a test set of 673 conflicts, with 137 unique \textit{Location--Insurgent} pairs throughout the whole time span (many pairs appear several times in different years). In total, it mentions 52 locations (with \textit{India} being the most frequent) and 128 armed insurgent groups (with \textit{ULFA} or \textit{United Liberation Front of Assam} being the most frequent). This test set is available for subsequent reuse (\url{http://ltr.uio.no/~andreku/armedconflicts/}).

\section{Predicting armed conflict participants}\label{sec:approach}
In this section, we provide a detailed description of our approach, starting with a synchronic example in \ref{subsec:syn} and then moving on to a toy diachronic example on one pair of years in \ref{subsec:dia}. In the next section \ref{sec:eval}, we conduct evaluation on the full test set.

\subsection{Synchronic modeling}\label{subsec:syn}
We first conducted preliminary experiments to assess the hypothesis that the embeddings contain semantic relationships of the type `insurgent participant of an armed conflict in the location'. To this end, we trained a CBOW model on the full English Gigaword corpus (about 4.8 billion tokens in total), with a symmetric context window of 5 words, vector size 300, 10 negative samples and 5 iterations. Words with a frequency less than 100 were ignored during training. We used Gensim \cite{Rehurek2010gensim} for training, and in terms of corpus pre-processing we performed lemmatization, PoS-tagging and NER using Stanford CoreNLP \cite{manning_corenlp}. Named entities were concatenated to one token (for example, \textit{United States} became \textit{United::States\_PROPN}). 

Then, we used the 137 \textit{Location--Insurgent} pairs derived in Section~\ref{sec:gold} to learn a projection matrix from the embeddings for locations to the embeddings for insurgents. The idea and the theory behind this approach are extensively described in \cite{mikolov2013translation} and \cite{kutuzovclustering}, but essentially it involves training a linear regression which minimizes the error in transforming one set of vectors into another. Finding the optimal transformation matrix amounts to solving $i$ normal equations (where $i$ is the vector size in the embedding model being used), as shown in Equation~\ref{eq:normalreg}:
\begin{equation}\label{eq:normalreg}
\boldsymbol\beta_i = (\textbf{X}^\intercal* \textbf{X} + \lambda * L)^{-1} * \textbf{X}^\intercal * y_i
\end{equation}
where $\textbf{X}$ is the matrix of 137 location word vectors (input), $y_i$ is the array of the $i$th components of 137 corresponding insurgent word vectors (correct predictions), $L$ is the identity matrix of the size $i$, with 0 at the top left cell, and $\lambda$ is a real number used to tune the influence of regularization term (if $\lambda = 0$, there is no regularization). $\boldsymbol\beta_i$ is the array of $i$ optimal coefficients which transform an arbitrary location vector into the $i$th component of the corresponding insurgent vector. After learning such an array for each vector component, we have a linear projection matrix which can `predict' an insurgent embedding from a location embedding.

To evaluate the resulting projections, we employed leave-one-out cross-validation, i.e., measuring the average accuracy of predictions on each pair from the test set, after training the matrix on all the pairs except the one used for the testing. The transformation matrix was dot-multiplied by the location vector from the test pair. Then, we found $n$ nearest neighbors in the word embedding model for this predicted vector. If the real insurgent in the test pair was present in these $n$ neighbors, the accuracy for this pair was 1, otherwise 0. In Table~\ref{tab:eval}, the average accuracies with different values of $\lambda$ and $n$ are reported.

\begin{table}
\begin{small}
\begin{tabular}{@{}lcccccccc@{}}
\toprule
&& \multicolumn{3}{c}{\textbf{loc$\rightarrow$group}}  && \multicolumn{3}{c}{\textbf{group$\rightarrow$loc}}\\ 
\cmidrule(lr){3-5}\cmidrule(l){7-9}
$\lambda$&&@1&@5&@10&&@1&@5&@10\\ 
\midrule
0.0 &&    0.0&     14.6 &    31.4 &	&    8.8 &  46.7 &\textbf{70.8}\\
0.5 &&    0.7& 19.0&\textbf{35.0}&	&    7.3 & 49.6 &70.1 \\
1.0 &&    \textbf{2.2} & \textbf{19.7}&  32.8 &	&   6.6 & 47.4 &66.4 \\ 
\bottomrule
\end{tabular}
\caption{Accuracies for synchronic projections from locations to armed groups, and vice versa}
\label{tab:eval}
\end{small}
\end{table}

The relations of this kind are not symmetric: it is much easier to predict the location based on the insurgent than vice versa (see the right part of Table~\ref{tab:eval}). Moreover, we find that the achieved results are roughly consistent with the performance of the same approach on the Google Analogies test set \cite{mikolov2013efficient}. We converted the semantic sections in the Analogies test set containing only nouns (\textit{capitals--common}, \textit{capitals--world}, \textit{cities in states}, \textit{currency} and \textit{family}) to sets of unique pairs. Then, linear projections with $\lambda=1.0$ were learned and evaluated for each of them. The average accuracies over these sections were 13.0@1, 48.77@5 and 62.96@10. 
 
The results on predicting armed groups are still worse than on the Google Analogies, because of 3 factors: 1) one-to-many relationships in the UCDP dataset (multiple armed groups can be active in the same location) make learning the transformation matrix more difficult; 2) the frequency of words denoting armed groups is lower than any of the words in the Google Analogies data set, thus, embeddings for them are of lower quality; 3) training the matrix on the whole Gigaword model is suboptimal, as the majority of armed groups were not active throughout all its time span.

All our experiments were also conducted using the very similar Continuous Skipgram models. However, as CBOW proved to consistently outperform Skipgram for our tasks, we only report results for CBOW, due to limited space.\footnote{It seems CBOW is often better than Skipgram with linear projections; cf. the same claim in \cite{kutuzovclustering}.}

To sum up this section, many-to-one semantic relations between locations and insurgents do exist in the word embedding models. They are less expressed than one-to-one relations like those in the Google Analogies test set, but still can be found using linear projections. In the next section, we trace the dynamics of these relations as the models are updated with new data.

\subsection{Diachronic modeling}\label{subsec:dia}
Our approach to using learned transformation matrices to trace armed conflict dynamics through time consists of the following. We first train a CBOW model on the subsection of Gigaword texts belonging to the year 1994. Then, we incrementally update (train) this same model with new texts, saving a new model after each subsequent year. The size of the yearly subcorpora is about 250--320 million content words each. Importantly, we also use vocabulary expansion: new words are added to the vocabulary of the model if their frequency in the new yearly data satisfy our minimal threshold of 15.\footnote{We did not experiment with different thresholds. It was initially set to the value which produced a reasonable vocabulary size of several hundred thousand words.} Each yearly training session is performed in 5 iterations, with linearly decreasing learning rate.  Note that we do not use any model alignment method (Procrustes, etc): our models are simply trained further with the new texts. A possible alternative to this can be incremental training of hierarchical softmax functions proposed in \cite{peng2017incrementally} or incremental negative sampling proposed in \cite{kaji2017incremental}; we leave it for future work.   

The experiment involves applying a learned transformation matrix across pairs of models. While in Section~\ref{sec:eval} we evaluate the approach across the entire Gigaword time period, this section reports a preliminary example experiment for the transition from 2000 to 2001 alone. This means we will have one model saved after sequential training for the years up to 2000, and one saved after year 2001. Our aim is to find out whether the \textit{Location--Insurgent} projection learned on the first model is able to reveal conflicts that appear in 2001. Thus, we extract from the UCDP dataset all the pairs related to the conflicts which took place between 1994 and 2000 (91 pairs total). The projection is trained on their embeddings from the first model (actually, on 79 pairs, as 12 armed group names were not present in the 2000 model and subsequently skipped). Then, this projection is applied to the second model embeddings of the 47 locations, which are subject to armed conflicts in the year 2001 (38 after skipping pairs with out-of-vocabulary elements). Table~\ref{tab:eval_dia1} demonstrates the resulting performance (reflecting how close the predicted vectors are to the actual armed groups active in this or that location). 

Note that out of 38 pairs from 2001, 31 were already present in the previous data set (ongoing conflicts). This explains why the evaluation on all the pairs gives high results. However, even for the new conflicts, the projection performance is encouraging. Among others, it managed to precisely spot the 2001 insurgency of the members of the \textit{Kosovo Liberation Army} in Macedonia, notwithstanding the fact that the initial set of training pairs did not mention Macedonia at all. Thus, it seems that the models at least partially `align' new data along the existing semantic axis trained before. 

\begin{table}
\center
\begin{small}
\begin{tabular}{@{}lccc@{}}
\toprule
Pairs (size)&@1&@5&@10\\ 
\midrule
All (38) &44.7 &73.7&84.2\\   
New (7) &14.3&28.6&42.9\\  
\bottomrule
\end{tabular}
\caption{Projection accuracy for the isolated example experiment mapping from 2000 $\rightarrow$ 2001.}
\label{tab:eval_dia1}
\end{small}
\end{table}

In the next section, we systematically evaluate our approach on the whole set of UCDP conflicts in the Gigaword years (1994--2010).

\section{Evaluation of diachronic models}
\label{sec:eval}

\begin{table*}[ht!]
\center
\begin{small}
\begin{tabular}{@{}p{55px}ccccccccccccc@{}}
\toprule
& \multicolumn{6}{c}{\textbf{Only in-vocabulary pairs}}&&\multicolumn{6}{c}{\textbf{All pairs, including OOV}} \\ 
\cmidrule(lr){2-7} \cmidrule(l){9-14}
& \multicolumn{3}{c}{\textbf{up-to-now}}&\multicolumn{3}{c}{\textbf{previous}}&&\multicolumn{3}{c}{\textbf{up-to-now}}&\multicolumn{3}{c}{\textbf{previous}}\\ 
\cmidrule(lr){2-4} \cmidrule(lr){5-7}\cmidrule(lr){9-11} \cmidrule(l){12-14}
               &@1   &@5    &@10  &@1   &@5   &@10  &&@1   &@5   &@10       &@1        &@5        &@10 \\
\cmidrule(lr){2-4} \cmidrule(lr){5-7}\cmidrule(lr){9-11} \cmidrule(l){12-14}
\textbf{Separate}       &0.0  &0.7   &2.1  &0.5  &1.1  &2.4  &&0.0  &0.5  &1.6       &0.4       &0.8       &1.8\\
\textbf{Cumulative}     &1.7  &8.3   &13.8 &2.9  &9.6  &15.2 &&1.5  &7.4  &12.2      &2.5       &8.5       &13.4\\ 
\textbf{Incr. static}   &54.9 &82.8  &90.1 &60.4&79.6 &84.8  &&20.8 &31.5 &34.2      &23.0      &30.3      &32.2\\ 
\textbf{Incr. dynamic}  &32.5 &64.5  &72.2 &42.6 &64.8 &71.5 &&28.1 &56.1 &\bf{62.9} &\bf{37.3} &\bf{56.7} &62.6     \\ 
\bottomrule
\end{tabular}
\caption{Average accuracies of predicting next-year insurgents on the basis of locations, using projections trained on the conflicts from all the preceding years (\textbf{up-to-now}) or the preceding year only (\textbf{previous}). Results for 3 baselines are shown along with the proposed \textbf{incremental dynamic} approach.}
\label{tab:eval_dia_inc_cum}
\end{small}
\end{table*}

To evaluate our approach on all the UCDP data, we again tested how good it is in predicting the future conflicts based on the projection matrices learned from the previous years. We did this for all the years between 1994 and 2010. The evaluation metrics are the same as in the Section~\ref{sec:approach}: we calculated the ratio of correctly predicted armed groups names from the conflict pairs, for which the UCDP datasets stated that these conflicts were active in this particular year. As before, the models employed in the experiment were incrementally trained on each successive year with vocabulary expansion. Words present in the gold standard but absent from the models under analysis were skipped. At the worst case, 25\% of pairs were skipped from the test set; on average, 13\% were skipped each year (but see the note below about the \textbf{incr. static} baseline). At test time, all the entities were lowercased.

We employ 3 baselines: 1) yearly models trained separately from scratch on the corpora containing texts from each year only (referred to as \textbf{separate} hereafter); 2) yearly models trained from scratch on all the texts from the particular year and the previous years (\textbf{cumulative} hereafter); 3) incrementally trained models without vocabulary expansion (\textbf{incr. static} hereafter).

Initially, the linear projections for all models were trained on all the conflict pairs from the past and present years, similar to Section~\ref{subsec:dia} (dubbed \textbf{up-to-now} hereafter). However, the information about conflicts having ended several years before might not be strongly expressed in the model after it was incrementally updated with the data from all the subsequent years. For example, the 2005 model hardly contains much knowledge about the conflict relations between Mexico and the \textit{Popular Revolutionary Army (EPR)} which stopped its activities after 1996. 
Thus, we additionally conducted a similar experiment, but this time the projections were learned only on the salient pairs (dubbed \textbf{previous}): that is, the pairs active in the last year up to which the model was trained. 

Table~\ref{tab:eval_dia_inc_cum} presents the results for these experiments, as well as baselines (averaged across 15 years). For the proposed \textbf{incr. dynamic} approach, the performance of the \textbf{previous} projections is comparable to that of the \textbf{up-to-now} projections on the accuracies @5 and @10, and is even higher on the accuracy @1 (statistically significant with \textit{t-test}, $p < 0.01$). Thus, the single-year projections are somewhat more `focused', while taking much less time to learn, because of less training pairs. 

The fact that our models were incrementally updated, not trained from scratch, is crucial. The results of the \textbf{separate} baseline look more like random jitter. The \textbf{cumulative} baseline results are slightly better, probably simply because they are trained on more data. However, they still perform much worse than  the models trained using incremental updates. This is because the former models are not connected to each other, and thus are initialized with a different layout of words in the vector space. This gives rise to formally different directions of semantic relations in each yearly model (the relations themselves are still there, but they are rotated and scaled differently).

The results for the \textbf{incr. static} baseline, when tested only on the words present in the test model vocabulary (the left part of the table), seem better than those of the proposed \textbf{incr. dynamic} approach. This stems from the fact that incremental updating with static vocabulary means that we never add new words to the models; thus, they contain only the vocabulary learned from the 1994 texts. The result is that at test time we skip many more pairs than with the other approaches (about 62\% in average). Subsequently, the projections are tested only on a minor part of the test sets.

Of course, skipping large parts of the data would be a major drawback for any realistic application, so the \textbf{incr. static} baseline is not really plausible. For comparison, the right part of Table~\ref{tab:eval_dia_inc_cum} provides the accuracies for the setup in which all the pairs are evaluated (for pairs with OOV words the accuracy is always 0). Other tested approaches are not much affected by this change, but for \textbf{incr. static} the performance drops drastically. As a result, for the all pairs scenario, incremental updating with vocabulary expansion outperforms all the baselines (the differences are statistically significant with \textit{t-test}, $p < 0.01$).

\section{Conclusion}
We have here shown how incrementally updated word embedding models with vocabulary expansion and linear projection matrices are able to trace the dynamics of subtle semantic relations over time. We applied this approach to the task of predicting armed groups active in particular geographical locations and showed that it significantly outperforms the baselines. However, it can be used for any kind of semantic relations. We believe that studying temporal shifts of such projections can lead to interesting findings far beyond the usual example of `king is to queen as man is to woman'.

To our best knowledge, the behavior of semantic relations in updated word embedding models was not explored before. Our experiments show that the models do preserve these `directions' and that the learned projections not only hold for the word pairs known to the initial model, but can also be used to predict relations for the new words. 

In terms of future work, we plan to trace how quickly incremental updates to the model `dilute' the projections, rendering them useless with time. We observed this performance drop in our experiments, and it would be interesting to know more about the regularities governing this deterioration. Also, for the particular task of analyzing armed conflicts, we plan to research ways of improving accuracy in predicting completely new armed groups not present in the training data, and the methods of filtering out locations not involved in armed conflicts. 

\bibliography{dia}
\bibliographystyle{emnlp_natbib}

\end{document}